\documentclass[conference]{IEEEtran}

\RequirePackage[l2tabu, orthodox]{nag}

\usepackage{cite}
\usepackage[T1]{fontenc}
\usepackage{amsmath,amssymb,amsfonts}
\usepackage{tcolorbox}

\ifCLASSOPTIONcompsoc
\usepackage[caption=false, font=normalsize, labelfont=sf, textfont=sf]{subfig}
\else
\usepackage[caption=false, font=footnotesize]{subfig}
\fi

\usepackage{booktabs, makecell}

\usepackage{xspace}
\makeatletter
\DeclareRobustCommand\onedot{\futurelet\@let@token\@onedot}
\def\@onedot{\ifx\@let@token.\else.\null\fi\xspace}

\makeatother

\usepackage{hyperref}
\newtcolorbox[auto counter]{pabox}[2][]{
title=Box~\thetcbcounter: #2,#1}

\begin{document}

\title{Natural Language Interaction with a Household Electricity Knowledge-based Digital Twin}


\author{\IEEEauthorblockN{%
     Carolina Fortuna\IEEEauthorrefmark{1}, Vid Han\v{z}el\IEEEauthorrefmark{1}, Bla\v{z}~Bertalani\v{c}\IEEEauthorrefmark{1}
}\IEEEauthorblockA{%
    \IEEEauthorrefmark{1}Department of Communication Systems, Jo\v{z}ef Stefan Institute, Slovenia\\
}
\{carolina.fortuna, blaz.bertalanic\}@ijs.si, vid.hanzel@gmail.com%
\\
}

\maketitle

\begin{abstract}
Domain specific digital twins, representing a digital replica of various segments of the smart grid, are foreseen as able to model, simulate, and control the respective segments. At the same time, knowledge-based digital twins, coupled with AI, may also empower humans to understand aspects of the system through natural language interaction in view of planning and policy making. This paper is the first to assess and report on the potential of Retrieval Augmented Generation (RAG) question answers related to household electrical energy measurement aspects leveraging a knowledge-based energy digital twin. Relying on the recently published electricity consumption knowledge graph that actually represents a knowledge-based digital twin, we study the capabilities of ChatGPT, Gemini and Llama in answering electricity related questions. Furthermore, we compare the answers with the ones generated through a RAG techniques that leverages an existing electricity knowledge-based digital twin. Our findings illustrate that the RAG approach not only reduces the incidence of incorrect information typically generated by LLMs but also significantly improves the quality of the output by grounding responses in verifiable data. This paper details our methodology, presents a comparative analysis of responses with and without RAG, and discusses the implications of our findings for future applications of AI in specialized sectors like energy data analysis.
\end{abstract}


\section{Introduction}

Due to growing population and technological advances, global electricity consumption, and consequently also CO\textsubscript{2} emissions are increasing~\cite{IEA2021elec}. The residential sector makes up 25\% of global electricity consumption and has great potential to increase efficiency and reduce CO\textsubscript{2} footprint without sacrificing comfort~\cite{IEA2021sector}. Stakeholders such as government and regulatory bodies, electricity system operators as well as individual household are increasingly relying on data for day to day operations and decision making~\cite{Darby2006, EPRI2009, FangSG}. Domain specific digital twins, representing a digital replica of various segments of the smart grid, are foreseen as able  to  model, simulate, and control the respective segments \cite{YU2022112407}. As early as 2011 \cite{tuegel2011reengineering}, digital twins were considered for aircraft structural life prediction while after 2014, the scientific community intensified digital twin related research especially in the manufacturing sector \cite{TAO2022372}. Generally, a digital twin is seen as a digital replica of a physical object or system and the vast majority of research assumes that it contains a simulation/modelling component. The authors of \cite{TAO2022372} identified geometric, physics, behaviour and rule model construction that were considered in the vast majority of the literature at the time.   

However, there are variations of digital twins, some of them not necessarily requiring a mathematical or geometrical model. For instance, the authors of \cite{van2020taxonomy} also found a definition according to which a digital twin "consists of three parts: physical product, virtual product, and
connected data that tie the physical and virtual product". Furthermore, digital twins are considered too slow to use in real-time scenarios, therefore a variation also refereed as digital shadow has been considered \cite{10.1007/978-3-030-49435-3_5}. They have been inspired by database views "named query on a database which can be reused in other queries or applications". Following this line of thought, we consider that a knowledge graph (i.e. Wikipedia) can represent a more structured and better connected database while natural language queries on it could be used for various applications. A universal digital twin represented as a knowledge graph has already been introduced in \cite{zhou2020knowledge}.

In this paper, we argue that a knowledge-based digital twin \cite{zhou2020knowledge} consisting of both high-level and specific energy domain knowledge and facts  represented in a structured machine readable format (i.e. knowledge graph) together with a large language model (LLM) may  empower humans to understand aspects of the electricity system through natural language interaction in a reliable and intuitive manner while avoiding LLM  hallucinations and deep domain and specific system  knowledge. In other words, the electricity knowledge graph introduced in \cite{hanzel2024kg} could be seen as a digital shadow of household electricity characteristics, which, together with the LLM forms a knowledge-based digital twin. \textit{The paper provides a view from the user end, and dependent on data, rather than at the physics end where the physical laws are the basis of the model}.


The first studies exploring the capabilities of LLMs and chatbots in the electric energy sectors are coming out \cite{dong2024exploring}, emphasizing opportunities and identifying challenges such as scarcity of domain-specific data in the pre-training of LLMs. At the same time, a household electricity knowledge graph has been developed and openly released \cite{hanzel2024kg}. However, to date, investigations on the potential enhancement of LLM responses via RAG using electrical energy data is missing. 

\begin{figure}[h]
    \centering
    \includegraphics[width=\linewidth]{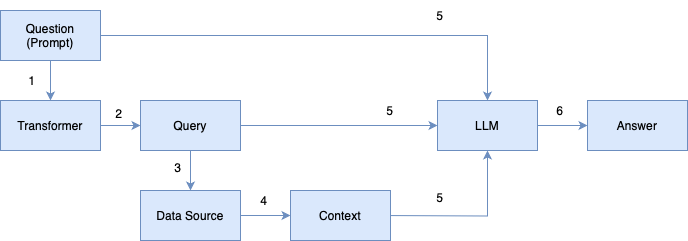}
    \caption{High level overview of the RAG process.}
    \label{fig:rag}
 \end{figure}
This paper is the first to assess and report on the potential of RAG-based question answers related to household electrical energy measurement aspects leveraging a knowledge-based energy digital twin. Relying on the recently published electricity consumption knowledge graph that actually represents a knowledge-based digital twin \cite{akroyd2021universal}, 
 we study the capabilities of ChatGPT, Gemini and Llama in answering electricity related questions. Furthermore, we compare the answers with the ones generated through a RAG techniques that leverages an existing electricity knowledge-based digital twin. We find that RAG helps with more precise responses in most of the cases. In other cases, it ignores some of the factual information possibly due to the bias build in during training while in other cases it may provide irrelevant answers.

The paper is organized as follows. Section \ref{sec:rag} provides background of LLMs and RAG, Section \ref{sec:rag_kg} discusses knowledge-based digital twin supported RAG for electricity while Section \ref{sec:example} illustrates the concrete approach employed for this study. Section \ref{sec:llm-rag} provides the analysis while Section \ref{sec:concl} concludes the paper.


\section{Background on LLMs and Retrieval Augmented Generation}
\label{sec:rag}
With Large Language Models (LLMs) disrupting existing established work processes promising increased efficiency, we expect their adoption also in supporting decision makers in the smart grids community. However, in some cases, LLMs are not reliable because they can generate plausible-sounding but incorrect or nonsensical answers due to their reliance on patterns in training data rather than true understanding. They are prone to "hallucinating" information, where they produce details that are not based on factual data, and they can also struggle with ambiguous or context-dependent queries~\cite{YangLLM}. Additionally, their outputs can reflect biases present in their training data, leading to potentially harmful or biased responses~\cite{beukeboom2019stereotypes}. Finally, their capabilities are enabled by their training data, encountering limitation with unseen private sources.

\subsection{Limitations of LLMs}
The limitations, including inherent uncertainty and variability make LLMs less dependable for applications requiring high accuracy and factual consistency such as data informed electricity related policy making and regulations. To mitigate for such cases, it is possible to 1) fine tune the LLMs on some specific data to adapt it to a particular task or domain~\cite{liu2024domain}, 2) adopt a two phase process involving context retrieval from external sources (such as  databases or documents) in response to a query and uses this information to generate a more accurate and contextually relevant answer~\cite{wu2024stark} or 3) prompt engineering by crafting and refining the inputs (prompts) given to a language model to elicit the best possible responses~\cite{Marvin2024}.

Each of the three methods is best suited for different scenarios. LLM fine-tuning is ideal for applications that demand high precision and customization in specific domains, such as specialized customer service or technical advice in fields like healthcare or finance~\cite{liu2024domain}. Prompt engineering is particularly useful for quickly improving model outputs across various tasks without the need for further training, making it suitable for content creation or rapid prototyping~\cite{Marvin2024}. Relying on external data sources, also referred to as retrieval augmented generation (RAG) is especially advantageous for tasks that require current and detailed information, such as answering factual questions or generating comprehensive reports, as it dynamically integrates the latest data into the model’s responses~\cite{wu2024stark}. Knowledge based digital twins \cite{akroyd2021universal} that contain structured domain specific knowledge in a form of a knowledge graph may be particularly suitable candidates for complementing LLMs in precise and human friendly natural language integration with large amount of data and information.
\begin{figure*}[!htpb]
    \centering
    \includegraphics[width=0.66\linewidth]{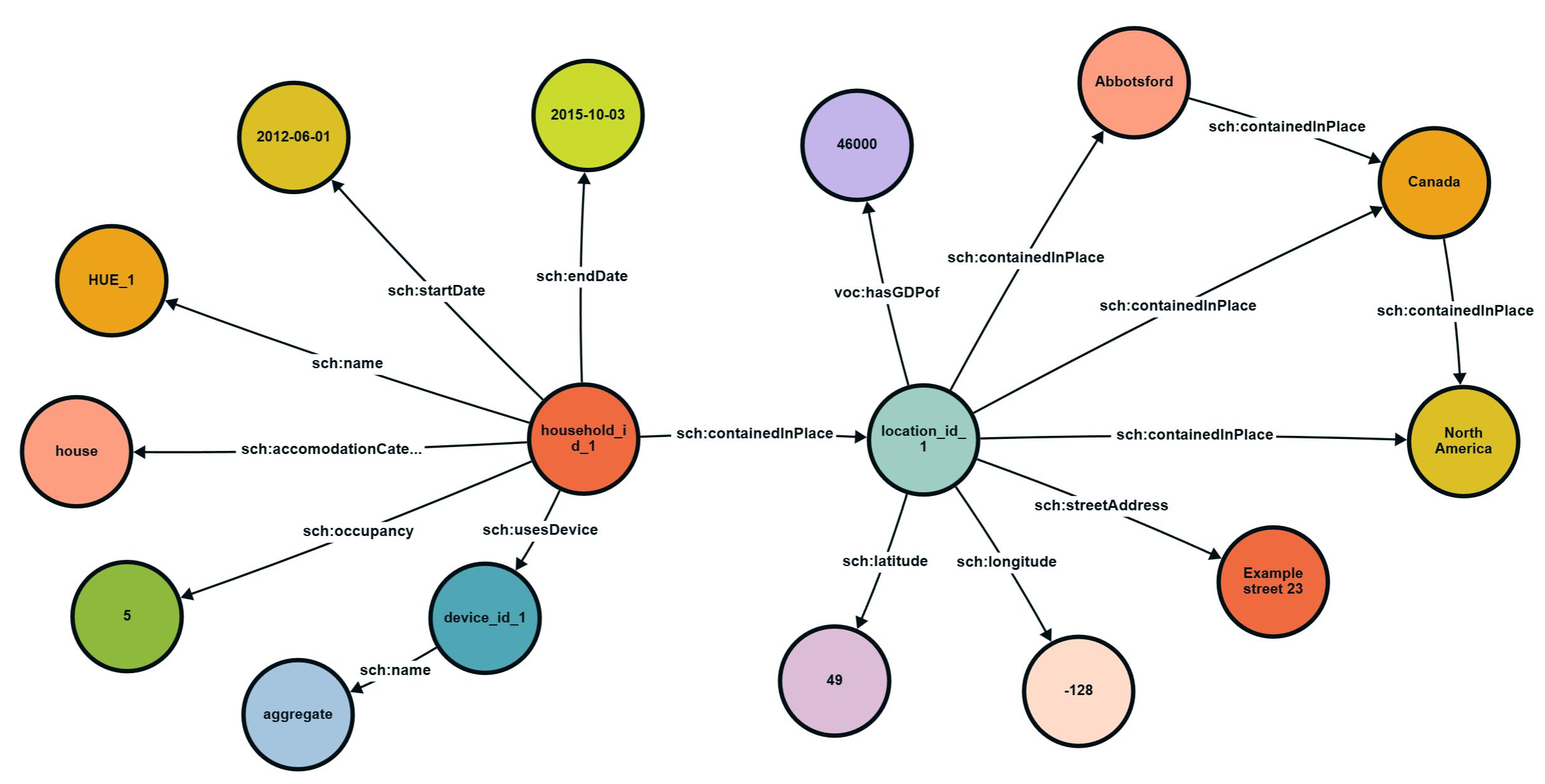}
    \caption{Household
Electricity Knowledge-based Digital Twin  Visualization~\cite{hanzel2024kg}.}
    \label{fig:kg}
 \end{figure*}
\subsection{Background on RAG}
Retrieval Augmented Generation (RAG) is a technique that combines the capabilities of retrieval-based and generation-based models to improve information generation and accuracy. As depicted in Figure \ref{fig:rag}, in RAG, the LLM produces and answer based on three elements: a question, a query and context. The question is typically provided by a human in natural language. It is mapped by a transformation block to a query as depicted by steps 1 and 2 in Figure \ref{fig:rag}. The transformation step may embed the question and the embedding subsequently represents the query. It can also transform the question from natural language to an alternative query language such as SQL or SPARQL. The query resulting from the transformation block is then sent to a retrieval system that fetches relevant documents or pieces of information from a large dataset as represented by steps 3 and 4 in the figure. This system can use a vector database when the query is an embedding, and relational database in case of an SQL query or a knowledge graph in the case of a SPARQL query. The information retrieved by the system (i.e. embeddings, records, triples) is referred to as context that is send together with the question and query to the LLM that generates the answer as illustrated in steps 5 and 6. With this process, the generative model produces a coherent and contextually accurate response. This approach leverages the extensive factual knowledge in the retrieval database and the linguistic and contextual abilities of generative models.

In general, RAG is considered useful because it enhances the accuracy and reliability of generated content by grounding it in real, retrieved data, reducing the likelihood of generating hallucinations or incorrect information \cite{wu2024stark}. This makes RAG particularly valuable in applications requiring precise and factual responses, such as question answering, customer support, and content creation, where both relevance and correctness are critical. By combining retrieval and generation, RAG achieves a more robust and informed response generation process.

\section{RAG with an Electricity Knowledge-based Digital Twin}
\label{sec:rag_kg}

To better understand the potential of RAG in enabling more accurate conversation for energy stakeholders, we study a case using an electricity knowledge-based digital twin embodied through a recently proposed knowledge graph~\cite{hanzel2024kg}.
With this approach, households, their properties and consumption details are modelled as a graph. As exemplified in Figure~\ref{fig:kg}, the nodes represent entities such as households, appliances, and locations, while edges represent the relationships between these entities. This structured model allows querying and machine reasoning over the household data, allowing extraction of patterns and insights.

 The graph model comprises three primary classes as depicted in Figure~\ref{fig:kg}: the household, its location, and its appliances which are interconnected. Each household node is connected to its respective location node and to all appliance nodes which are contained within the household. The household class describes the household with the start and end date of electricity consumption measurements, occupancy, square footage of the living space, and dwelling type (e.g., house or apartment). The appliance class contains properties detailing appliance electricity consumption such as average daily consumption and average consumption during an on/off event. The location class describes the household location with its coordinates along with the city, country, and continent information. The location class integrates with WikiData and DBpedia to enrich the city and country information. Additionally, it includes demographic properties such as GDP, average wage, population density, energy prices, and the carbon intensity of electricity production. The KG can be explored further on it's LodView deployment~\footnote{\url{https://elkg.ijs.si/resource/public-households/REFIT_1}}.
The KG is encoded in RDF, stored in a Blazegraph database and can be queried via a SPARQL endpoint. So we assume that the transformation block from Figure \ref{fig:rag} transforms the natural language to the corresponding SPARQL query, and that the LLM receives the natural language question, the SPARQL query and the triples that are retrieved from the KG. We evaluate the RAG with two versions of ChatGPT (4o and 4), Gemini 1.5 and Llama-3-8b-chat-hf. 

Based on the knowledge available in the twin we select questions that require increasing levels of precision to study the potential of the proposed concept as follows:

\begin{itemize}
    \item \textit{Prompt 1} "Enumerate in one short sentence the electricity consumption datasets collected in the UK." (For Gemini it returned nothing so we rephrased to: "How many electricity consumption datasets were collected in the UK?")
    \item \textit{Prompt 2} "Enumerate in one short sentence the available electricity datasets located in countries with a GDP per capita higher than \$50000."  
    \item \textit{Prompt 3} "Enumerate in one short sentence the available electricity datasets that are not located in Europe and are located in a place with a high education level."
    \item \textit{Prompt 4} "Can you explain the load profile of house 1 in the REFIT dataset?"
\end{itemize}

\subsection{ChatGPT 4o and RAG Example}
\label{sec:example}
To better understand the methodology employed in this paper, we provide an example with ChatGPT4o. Given Prompt 1, we first use the regular non-RAG approach followed by the RAG approach as listed in the boxes below. It can be seen that the response to the natural language question returns five datasets, i.e. NEED, ECUK, Smart Meter Energy Consumption Data, and a pointer to a portal. The response with the RAG returns the list of datasets available in the electricity KG. Only one of the datasets, namely REFIT, overlap between the two answers.

\section{LLM and RAG Analysis}
\label{sec:llm-rag}

In this section we summarize and analyze the responses provided by the selected LLMs to the four questions without employing RAG and with RAG.

\subsection{Analysis related to Prompt 1}

Ther answers to Prompt 1 are summarized in Table \ref{tab:q1} and the non RAG prompt can be seen in Box~\ref{box:p1_no_rag} and the RAG prompt in Box~\ref{box:p1_gpt4_rag}. It can be seen from the first row of the table and from Box~\ref{box:p1_gpt4o_no_rag_answer} that ChatGPT 4o provided 6 datasets and one data repository as an answer without employing RAG and we can see in Box~\ref{box:p1_chatgpt4_rag} that with RAG it returns 3 datasets as there are only 3 UK datasets present in the KG. There is one overlap in the found datasets with the two approaches. Overall, it provided the most precise answers given the natural language question. The answer~\ref{box:p1_chatgpt4_rag} of ChatGPT 4 was of equal quality.

\begin{table*}[h]
  \centering
  \caption{Quality of the answers to Prompt 1: "Enumerate in one short sentence the electricity consumption datasets collected in the UK." (For Gemini it returned nothing so we rephrased to: "How many electricity consumption datasets were collected in the UK?").}
  \label{tab:q1}
  \begin{tabular}[width=0.5\linewidth]{lllll}
    \toprule
    LLM & non-RAG Answer & RAG Answer & Overlap & Comments\\
    \midrule
    ChatGPT 4o & 6 datasets, 1 repo & 3 datasets & 1 dataset &\\\midrule
    ChatGPT 4 & 6 datasets, 1 repo & 3 datasets & 1 dataset & Lengthier than solicited textual response.\\\midrule
    Llama-3-8b-chat-hf & 0 datasets & 3 datasets & 0 datasets & Sources that collected data rather than datasets.\\\midrule
    Gemini & 1 dataset and 2 repos & 3 datasets & 0 datasets & Gives examples and pointers, avoids being precise.\\
  \bottomrule
\end{tabular}
\end{table*}

\begin{pabox}[colframe=blue!75!black, colback=blue!10, label={box:p1_no_rag}]{\textit{Prompt 1} non RAG}
\footnotesize
Enumerate in one short sentence the electricity consumption datasets collected in the UK?
\end{pabox}
\begin{pabox}[colframe=green!75!black, colback=green!10, label={box:p1_gpt4o_no_rag_answer}]{ChatGPT 4o non-RAG answer to Prompt 1}
\footnotesize
In the UK, key electricity consumption datasets include the National Energy Efficiency Data-Framework (NEED), the annually updated Energy Consumption in the UK (ECUK) datasets, real-time Smart Meter Energy Consumption Data, regional and local Sub-National Electricity Consumption Data, detailed data from the Household Electricity Survey, diverse datasets from the National Grid Electricity System Operator (ESO) Data Portal, and the REFIT Smart Home Dataset.
\end{pabox}
\begin{pabox}[colframe=blue!75!black, colback=blue!10, label={box:p1_gpt4_rag}]{\textit{Prompt 1} with RAG }
\footnotesize
\textit{Question}:\\
Enumerate in one short sentence the electricity consumption datasets collected in the UK?\\

\textit{Query}:\\

PREFIX voc: <https://elkg.ijs.si/ontology/> \\
PREFIX saref: <https://saref.etsi.org/core/> \\
PREFIX rdf: <http://www.w3.org/1999/02/22-rdf-syntax-ns> \\
PREFIX schema: <https://schema.org/> \\
SELECT DISTINCT ?prefix ?countryName  WHERE \{ \\
 ?house rdf:type schema:House . \\
?house schema:name ?houseName . \\
?house schema:containedInPlace ?place . \\
?place schema:containedInPlace ?country . \\
?country rdf:type schema:Country . \\
?country schema:name ?countryName . \\
FILTER(?countryName = "United Kingdom") . \\
BIND(STRBEFORE(?houseName, "\_" ) as ?prefix) . \\
\} \\

\textit{Enhanced Context}:\\
prefix	countryName \\
IDEAL	United Kingdom \\
REFIT	United Kingdom \\
UKDALE	United Kingdom \\
\end{pabox}

\begin{pabox}[colframe=green!75!black, colback=green!10, label={box:p1_gpt4o_rag}]{ChatGPT 4o RAG answer to Prompt 1}
    \footnotesize
    The electricity consumption datasets collected in the UK include IDEAL, REFIT, and UKDALE. 
    
\end{pabox}

    \begin{pabox}[colframe=green!75!black, colback=green!10, label={box:p1_chatgpt4_rag}]{ChatGPT 4 Answer to Prompt 1 with RAG}
    \footnotesize
   The electricity consumption datasets collected in the UK include
IDEAL, REFIT, and UKDALE
    \end{pabox}
As can be seen in the 3rd row of Table \ref{tab:q1}, Llama-3-8b-chat-hf does not provide any datasets when not using RAG. Instead it provides sources that collected data as can be seen from the answer in Box~\ref{box:llama_1_no_rag}. When using RAG, Llama-3-8b-chat-hf lists all of the datasets provided as context. 
For the case of Gemini when not using RAG the chatbot avoids being specific in its answers. It however gives examples and references to where to search more as illustrated in Box~\ref{box:p1_gemini_no_rag}. When using RAG it states that there are atleast 3 datasets collected in the UK based on the enhanced context and list the 3 datasets present in the enhanced context as can be seen in Box~\ref{box:p1_gemini_rag}.


\begin{pabox}[colframe=green!75!black, colback=green!10, label={box:p1_gemini_no_rag}]{Gemini non-RAG answer to Prompt 1}
\footnotesize
  There isn't a single definitive source that provides a count of all electricity consumption datasets collected in the UK. However, several resources indicate the existence of multiple datasets:

The UKERC Energy Data Centre catalog lists datasets related to electricity consumption [1]. While it doesn't provide a total count, it highlights various datasets.

A sample of anonymized electricity consumption data is available, implying there's a larger dataset used for national statistics [2].

A dataset on Kaggle provides information on electricity demand in Great Britain, suggesting collection of electricity consumption data [3].

These examples showcase multiple datasets on electricity consumption in the UK, but not a specific count.
\end{pabox}

\begin{pabox}[colframe=green!75!black, colback=green!10, label={box:llama_1_no_rag}]{Llama-3-8b-chat-hf non-RAG Answer to Prompt 1}
\footnotesize
 The UK has collected electricity consumption datasets from various sources, including National Grid, BEIS, OpenEI, UK Power Networks, ONS, and research institutions, covering different time periods, geographic areas, and data granularity.
\end{pabox}

\begin{pabox}[colframe=green!75!black, colback=green!10, label={box:p1_gemini_rag}]{Gemini RAG answer to Prompt 1}
\footnotesize
The number of electricity consumption datasets collected in the UK is at least 3, based on the provided enhanced context.

Here's the breakdown: \\
IDEAL\\
REFIT\\
UKDALE\\
\end{pabox}

\begin{table*}[!htp]
  \centering
  \caption{Quality of the answers to Prompt 2: "Enumerate in one short sentence the available electricity datasets located in countries with a GDP per capita higher than \$50000."}
  \label{tab:q2}
  \begin{tabular}[width=\linewidth]{lllll}
    \toprule
    LLM & non-RAG Answer & RAG Answer & Overlap & Comments\\
    \midrule
    ChatGPT 4o & 2 datasets & 5 datasets & 1 dataset & REFIT, ECO\\\midrule
    ChatGPT 4 & 3 datasets & 5 datasets & 1 dataset & Pecan, REDD and GREEND\\\midrule
    Llama-3-8b-chat-hf & 0 datasets & 5 datasets & 0 datasets & Countries with 50k GDP and their agencies that collect data.\\\midrule
    Gemini & 0 datasets & 0 datasets & 0 datasets & Gives examples and pointers, avoids being precise, finds \\ & & & &  problems in the SPARQL query.\\
  \bottomrule
\end{tabular}
\end{table*}

\subsection{Analysis related to Prompt 2}

The answers to Prompt 2 are summarized in Table \ref{tab:q2}. It can be seen from the first row of the table that ChatGPT 4o provided 2 datasets, i.e. REFIT and ECO, without employing RAG and 5 datasets by employing RAG. There is one overlap in the found datasets with the two approaches. The answer of ChatGPT 4 was of 3 datasets, i.e. Pecan, REDD and GREEND without RAG and 5 datasets with RAG and with the REDD dataset overlapping. It can be seen that the answers of ChatGPT 4o and ChatGPT 4 vary and this variation is due to the fact that country GDPs have changed between the training time of the two models. The UK, where REFIT was collected, used to have below 50k \$ GDP in the past while this has recently changed according to the IMF \footnote{\url{https://www.imf.org/external/datamapper/NGDPDPC@WEO/GBR/DEU/GRC}}. We can conclude that the answers with RAG are more predictable as it could retrieve examples with GDP at the time of collection, last year or now. However, when such tools are integrated into the standard enterprise business processes, safeguards that control for untimeely information should be considered.

\begin{pabox}[colframe=green!75!black, colback=green!10, label={box:p2_llama_no_rag}]{Llama-3-8b-chat-hf non-RAG answer to Prompt 2}
\footnotesize
Here are some available electricity datasets located in countries with a GDP per capita higher than \$50,000:\\
United States: Energy Information Administration (EIA) datasets on electricity generation, consumption, and prices\\
Canada: Natural Resources Canada datasets on electricity generation, consumption, and prices\\
.....

Note that this is not an exhaustive list, and there may be other datasets available in these countries. Additionally, some datasets may be available through international organizations such as the International Energy Agency (IEA) or the Organization for Economic Cooperation and Development (OECD).
\end{pabox}

As can be seen in the third line of Table \ref{tab:q2}, Llama doesn't return any dataset in the non-RAG approach, however as can also be seen in Box~\ref{box:p2_llama_no_rag}, it finds countries that respect the GDP constraint and provides the agencies in the respective countries that collect such data as can be seen in the box below. With RAG, the answer is consistent with the datasets provided as context. For this prompt, Gemini does not identify any datasets in either approach as can be seen from the last row of the table. It provides a generic answer in the non-RAG version and finds issues in the SPARQL query in the RAG version.

\begin{table*}[h]
  \centering  
  \caption{Quality of the answers to Prompt 3: "Enumerate in one short sentence the available electricity datasets that are located in Europe and had at the time of recording an electricity price higher than 0.25€/kWh."}
  \label{tab:q3}
  \begin{tabular}[width=\linewidth]{lllll}
    \toprule
    LLM & non-RAG Answer & RAG Answer & Overlap & Comments\\
    \midrule
    ChatGPT 4o & 1 dataset & 10 datasets & 1 dataset & ECO\\\midrule
    ChatGPT 4 & 2 datasets & 10 datasets & 0 dataset & Pricing datasets, not measurement.\\\midrule
    Llama-3-8b-chat-hf & 0 datasets & 8 datasets & 0 datasets & List of national TSO having wholesale prices.\\\midrule
    Gemini & 0 datasets & 7 datasets & 0 datasets & Non-RAG provides a statement that doesn't make sense. \\
  \bottomrule
\end{tabular}
\end{table*}

\subsection{Analysis related to Prompt 3}

The answers to Prompt 3 are summarized in Table \ref{tab:q3}. It can be noticed that the non-RAG version provides limited answers, even with the more capable ChatGPT versions in the first two rows. Furthermore, ChatGPT 4 refers to electricity pricing datasets rather than electricity datasets collected when the price was higher than 0.25€/kWh.  ChatGPT 4o captured the nuance and returns one correct dataset that overlaps with on the the datasets returned using RAG. The non-RAG answers with Llama and Gemini are not useful for this prompt indicating relatively shallower semantic capabilities. The answers generated with the RAG versions seem to omit the information provided as context. Further research may be required to understand whether  the more precise the question, the more it confuses the LLM that becomes unable to leverage the query and context in the response generation.

\subsection{Analysis related to Prompt 4}
The fourth prompt assesses the value of retrieving a load profile image from the KG for understanding the consumption behavior of that respective house. As expected, when asking ChatGPT 4o and 4 the following: "Can you explain the load profile of house 1 in the REFIT dataset?", it provides a well structured, but generic answer. It provides valuable guidelines on how to analyze the house 1 REFIT load profile as can be seen in Box~\ref{box:p4_gpt4_no_rag}. However, when adding the load profile image, extracted from the KG as can be seen in Box~\ref{box:p4_gpt_prompt}, both versions are able to provide specific analysis based on the numbers present in the image. However, as can be seen from the answer in Box~\ref{box:p4_gpt4_4o}, ChatGPT 4 made less factual mistakes interpreting the image compared to ChatGPT 4o. ChatGPT 4o identifies a drop between 1-6 hour after after a peak from 0 to 1 hour that is not present in the load profile. ChatGPT 4  extracts the morning peak 0-5 hours followed by the morning drop 5-10 hours that is more consistent with the profile.

\begin{pabox}[colframe=blue!75!black, colback=blue!10, label={box:p4_gpt_prompt}]{\textit{Prompt 4} with RAG}
\footnotesize
Can you explain the load profile of house 1 in the REFIT dataset? (added load profile plot for context)
\includegraphics[width=0.3\linewidth]{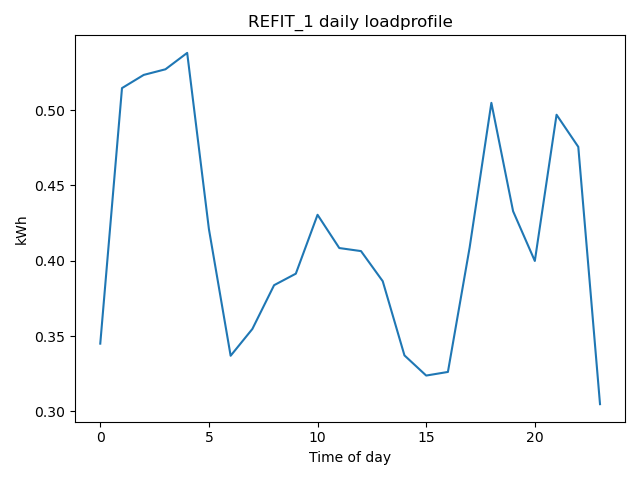}
\end{pabox}
\begin{pabox}[colframe=green!75!black, colback=green!10, label={box:p4_gpt4_4o}]{Example portion of factual answer ChatGPT 4o vs 4}
\footnotesize
ChatGPT 4o

1. \textbf{Early Morning Peak (Around 0 to 1 hour)}:

   - The load profile shows a significant peak in the early morning hours. This could be due to the use of appliances such as heating systems, or it might indicate the end of a high-consumption activity that started late in the previous evening.

2. \textbf{Drop After the Peak (Around 1 to 6 hours)}:

   - There is a noticeable drop in consumption after the early morning peak, reaching its lowest point just before the morning hours. This suggests a period when the household is mostly inactive, likely because occupants are asleep.

ChatGPT 4

1. \textbf{Early Morning Peak (0-5 Hours)}:
   - The profile starts with a high peak early in the morning. This could indicate overnight usage of appliances such as heaters, especially if this is a winter day profile, or other devices left running through the night.

2. \textbf{Morning Drop (5-10 Hours)}:
   - After the initial peak, there is a significant drop in consumption. This decrease might correspond to a period when the residents are sleeping and most active appliances are turned off, or it could reflect a period just before the household starts its daily activities.
\end{pabox}

\begin{pabox}[colframe=green!75!black, colback=green!10, label={box:p4_gpt4_no_rag}]{Well structured instructions provided by ChatGPT}
\footnotesize
The REFIT dataset is a detailed collection of electrical consumption data from several households in the UK, gathered as part of the REFIT (Retrofit) project. The load profile of a house in this dataset refers to the pattern of electrical usage over a specified period. Specifically for house 1 in the REFIT dataset, the load profile can be analyzed by looking at the time-series data of electrical consumption recorded.

Here's a step-by-step explanation of what the load profile might reveal:

\textbf{Data Collection}

1. \textit{Time Series Data}: The dataset contains time-stamped readings of electrical usage. Each entry typically records the power consumption (in watts or kilowatts) at regular intervals (e.g., every minute).

\textbf{Key Components of the Load Profile}

2. \textit{Daily and Weekly Patterns}:

   - \textit{Daily Variations}: The load profile can show how electrical consumption varies throughout the day. Peaks might correspond to times when occupants are home and using appliances (morning and evening), while troughs could indicate periods of inactivity (late night).
   
   - \textit{Weekly Variations}: Similarly, the profile might reveal differences between weekdays and weekends, reflecting different routines and usage patterns.

3. \textit{Seasonal Variations}:

   - Changes in consumption due to seasonal factors, such as increased heating in winter or air conditioning in summer.

.....

\textbf{Example Interpretation}

For house 1, the load profile might show:

- Morning peaks as occupants prepare for the day (e.g., 7-9 AM).

- Evening peaks when occupants return home and use appliances (e.g., 6-9 PM).

- Lower consumption during working hours if the house is unoccupied.

.....

If you have access to the actual data, we could plot and analyze specific examples to provide more detailed insights.
\end{pabox}

\section{Conclusions}
\label{sec:concl}
In this study, we explored the capabilities of advanced large language models (LLMs) like ChatGPT, Gemini, and Llama alongside the Retrieval Augmented Generation (RAG) approach using an electricity knowledge-based digital twin for answering complex questions about household electricity measurements and knowledge. Our analysis illustrated that while LLMs can provide quick and broadly accurate responses, integrating them with RAG, which pulls precise data from a specialized electricity knowledge-based digital twin, significantly enhances the precision and details available of the responses. This synergy between generative AI and targeted data retrieval proves especially beneficial in fields like energy data analysis, where precision and context-specificity are paramount. As such, RAG not only mitigates some common flaws in LLMs, such as the generation of plausible yet incorrect information, but also enriches the model's ability to handle specific, nuanced queries that are critical for data-driven decision-making in the energy sector. This integration represents a promising direction for further research and application, particularly in enhancing the reliability and utility of AI in specialized domains. However, a larger scale research effort on more domain specific data is needed to fully asses the robustness or RAG and the influence of potential build-in LLM bias that would disregard the context in the generated answer.
\section{Acknowledgments}
This work was funded in part by the Slovenian Research Agency under the grant P2-0016 and L2-50053. The authors would also like to acknowledge the anonymous reviewers providing valuable insights that helped improve the clarity of the paper.

\bibliographystyle{IEEEtran}
\bibliography{bibliography}

\end{document}